\documentclass[10pt,twocolumn,letterpaper]{article}

\usepackage[pagenumbers]{wacv} 

\usepackage[cmex10]{amsmath}

\usepackage{multirow}
\usepackage{makecell}
\usepackage[shortlabels]{enumitem}

\usepackage{comment}
\usepackage{booktabs}
\usepackage{multirow}
\usepackage{multicol}
\usepackage[table]{xcolor}
\usepackage{tabularx}
\usepackage{amssymb}
\usepackage{algorithm}
\usepackage{algpseudocode}
\usepackage{verbatim}
\usepackage{amsmath}

\DeclareMathOperator*{\argmin}{arg\,min}

\usepackage{tikz}
\usepackage[pagebackref,breaklinks,colorlinks,bookmarks=false]{hyperref}

\def\wacvPaperID{1090} 
\def\confName{WACV}
\def\confYear{2025}
\usepackage[capitalize]{cleveref}
\crefname{section}{Sec.}{Secs.}
\Crefname{section}{Section}{Sections}
\Crefname{table}{Table}{Tables}
\crefname{table}{Tab.}{Tabs.}

\begin{document}

\title{Meta-Learning for Color-to-Infrared Cross-Modal Style Transfer}

\author{Evelyn A. Stump\textsuperscript{1}
\and
Francesco Luzi\textsuperscript{1}
\and
Leslie M. Collins\textsuperscript{1}
\and
Jordan M. Malof\textsuperscript{2}
\and
\textsuperscript{1} Electrical and Computer Engineering, Duke University\\
\textsuperscript{2} Electrical Engineering and Computer Science, University of Missouri\\
{\tt\small \{evelyn.stump, francesco.luzi, leslie.collins\}@duke.edu},\\
{\tt\small jmdrp@missouri.edu}}

\def\eg{\emph{e.g}\bmvaOneDot}
\def\Eg{\emph{E.g}\bmvaOneDot}
\def\etal{\emph{et al}\bmvaOneDot}

\maketitle

\begin{abstract}
Recent object detection models for infrared (IR) imagery are based upon deep neural networks (DNNs) and require large amounts of labeled training imagery. However, publicly available datasets that can be used for such training are limited in their size and diversity.  To address this problem, we explore cross-modal style transfer (CMST) to leverage large and diverse color imagery datasets so that they can be used to train DNN-based IR image-based object detectors.  We evaluate six contemporary stylization methods on four publicly-available IR datasets – the first comparison of its kind - and find that CMST is highly effective for DNN-based detectors.   Surprisingly, we find that existing data-driven methods are outperformed by a simple grayscale stylization (an average of the color channels).  Our analysis reveals that existing data-driven methods are either too simplistic or introduce significant artifacts into the imagery.  To overcome these limitations, we propose meta-learning style transfer (MLST), which learns a stylization by composing and tuning well-behaved analytic functions.  We find that MLST leads to more complex stylizations without introducing significant image artifacts and achieves the best overall detector performance on our benchmark datasets.   
\end{abstract}

\vspace{-1.5em}
\section{Introduction}
\vspace{-.25em}
\label{sec:introduction}
Object detection is a computer vision task that has been dominated in recent years by deep neural networks (DNNs) \cite{jing2019neural}.  DNNs require large quantities of annotated data to train effectively \cite{jing2019neural}. Therefore, the success of DNNs can be partially attributed to the availability of large and public datasets, such as COCO \cite{lin2014microsoft}.  While such datasets are available for many color (RGB) imagery recognition problems, the datasets available for infrared (IR) imagery problems are far more limited.  Several public IR datasets now exist (e.g., CAMEL \cite{Gebhardt2019}, FLIR \cite{flirdataset}), however, they are smaller and less rich than color-based datasets, limiting the ability to effectively train or evaluate IR recognition models. There are many general strategies to mitigate this problem: more efficient use of data (e.g., data augmentation \cite{Shorten2019}, self-supervised learning \cite{Jaiswal2020}), more effective modeling (e.g., ResNet \cite{He2016}, Transformers \cite{wolf2020transformers}),  transfer learning \cite{weiss2016survey}, and more. 

\begin{figure}
\centering
\includegraphics[width=0.8\columnwidth]{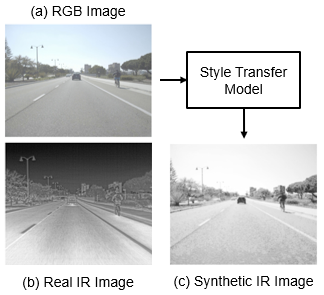}
\vspace{-0.75em}
\caption{An example of (a) an RGB image, (b) a real IR image and (c) a synthetic IR image produced by a style transfer model. Images from \cite{flirdataset}}

\label{fig:cmst_example}
\vspace{-1.75em}
\end{figure}

Another recently studied strategy to overcome training data limitations is RGB-to-IR cross-modal style transfer (CMST) \cite{Li2020}, which is the focus of this work.  The goal of CMST is to transform RGB (i.e., color) imagery so that it appears as though it were collected under similar conditions using an IR camera \cite{Li2020}. Fig. \ref{fig:cmst_example} illustrates CMST with a real-world pair of co-collected RGB and IR imagery. If an accurate RGB-to-IR transformation can be found, it may be possible to directly leverage large existing RGB datasets (e.g., COCO) to train IR object detection models, alleviating the current bottleneck of IR training data.  

Unfortunately, deriving an accurate CMST model is challenging. The IR image of a scene depends upon many factors, such as the material composition and solar insolation of the scene content, as well as recent weather conditions in the scene (e.g., ambient temperature).  Modeling the relationship between IR imagery and these underlying factors is challenging, even when full information about them is available.  However, in CMST we must infer an IR image of some scene using only an RGB image of that scene. As a result, CMST models do not have (direct) access to many factors that influence the appearance of the IR imagery, and there are many valid IR images that could be paired with a given input RGB image. Here each plausible IR image corresponds to a different setting of the underlying scene conditions (e.g., temperature, insolation).

\begin{figure}[tp]
\centering
\includegraphics[width=1.\columnwidth]{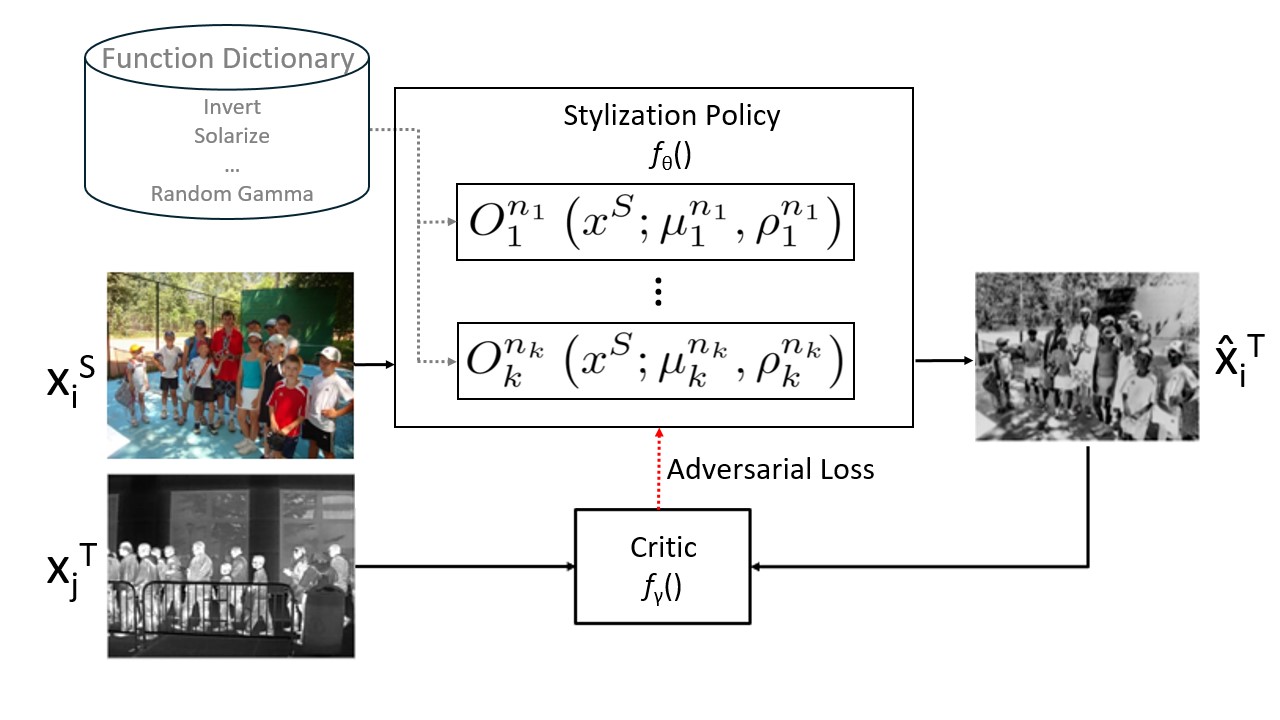}
\vspace{-0.60em}
\caption{ This figure shows a diagram of our proposed MLST model. RGB imagery is stylized by a composition of k functions ($O^{n}$) and function parameters ($\mu_k^n, \rho_k^n$) sampled from a distribution of $N$ possible function where $n \in [1 ,2, \cdots, N]$. $\mu_k^n$ and $\rho_k^n$ are sampled from a learned probability distribution where $\rho$ determines the probability of application for each function and $\mu$ is the function specific parameter. This composition of functions $f_\theta()$ is termed a Stylization Policy. The loss of an adversarial critic (red line) is used as the training signal for the policy. }

\label{fig:mlst_concept}
\vspace{-1.5em}
\end{figure}

\textbf{Existing Work on RGB-to-IR CMST.} Despite the aforementioned challenges of RGB-to-IR CMST, recently proposed CMST methods have shown promise \cite{Li2020}. 
For example, the authors in \cite{Herrmann2019} hand-crafted an effective CMST by composing several common image transformations (e.g., contrast enhancement, blurring).  However, due to the complexity of the RGB-to-IR CMST, another approach has been to use machine learning (ML) \cite{li2019multi, kniaz2018thermalgan, uddin2021converting, ozkanouglu2022infragan, lee2023edge} to \textit{infer} an effective model using data.  In this case, an ML model learns a function that maps RGB into IR imagery using examples of imagery from each modality.  \textit{In principle}, ML models can leverage the increasingly large datasets of RGB and IR imagery to learn more complex and accurate CMST models than have been achieved with hand-crafted approaches.

Despite the variety of RGB-to-IR CMST methods, one limitation of existing work is that there is yet to be a comprehensive comparison of approaches.  In particular, existing studies typically only compare their proposed methods against just one or two approaches. As a result, there are no studies exploring, for example, whether ML-based approaches outperform hand-crafted methods or simple baseline approaches like grayscale transformations.  Existing work also only compares competing CMST methods on just one or two datasets.  In recent years, several new datasets have become available, making it possible to conduct more rigorous comparisons of RGB-to-IR CMST methods. Without a rigorous comparison of methods, it is difficult to determine the absolute and relative performance of existing RGB-to-IR CMST approaches, and the extent of methodological progress over time.

\textbf{Contributions of This Work.} In this work we conduct a large-scale comparison of recent RGB-to-IR CMST models on four different publicly available IR datasets.  Our comparison includes several hand-crafted and data-driven CMST models \cite{Herrmann2019}, respectively, as well as simple baseline methods (e.g., identity and grayscale transformations). Additionally, we compare to the current state-of-the-art RGB-IR CMST methods: ThermalGAN \cite{kniaz2018thermalgan}, and InfraGAN \cite{ozkanouglu2022infragan}. CMST can be seen as a special case of general style transfer, and therefore we also include several successful general style transfer methods: WCT2 \cite{yoo2019photorealistic}, CycleGAN  \cite{zhu2017unpaired}, and CyCADA \cite{hoffman2018cycada}. We evaluate the quality of each CMST model by using its synthetic IR imagery to train IR object detection models, following recent CMST studies \cite{Kieu2020, Liu_2018_CVPR_Workshops, kniaz2018thermalgan, ozkanouglu2022infragan}.  Based upon our benchmark experiments we find that existing ML-based CMST methods are outperformed by manually chosen combinations of conventional transformations (e.g., grayscale, intensity inversions).  Motivated by these results, we propose meta learning style transfer (MLST), which attempts to automatically learn the best combinations of such transformations, making such approaches more data-driven. A diagram of MLST along with real stylized RGB→IR imagery is shown in Fig.  \ref{fig:mlst_concept}. We find that MLST learns more complex CMST models than hand-crafted models, but without altering the content of the underlying scene.

Our specific contributions can be summarized as follows: 
\begin{itemize}
\item \textit{The first comprehensive benchmark comparison of RGB-to-IR CMST models}. We present a performance comparison of numerous applicable approaches for RGB-to-IR CMST, including hand-crafted methods, general image stylization methods, and methods specifically designed for RGB-to-IR CMST. We evaluate the methods on four public benchmark datasets. 
\item \textit{Meta-Learning Style Transfer (MLST), a state-of-the-art RGB-to-IR CMST model}. MLST offers substantially greater performance on average compared with all other methods in our benchmark. We also present an analysis of MLST showing that it is learning effective RGB-to-IR models and closing the visual gap between RGB and IR imagery.    
\end{itemize}

The organization of this paper is as follows: Section \ref{sec:related_work} discusses related work. 
In Section \ref{sec:problem_setting} we precisely define the CMST problem setting mathematically. Section \ref{sec:mlst_description} presents MLST. 
In Section \ref{sec:baseline_methods} we discuss the baseline CMST models we study in this work and review what datasets are used. In Section \ref{sec:experiments} we present our experimental design and results; Section \ref{sec:analysis}  we perform additional experiments to analyze the primary benefits of MLST. In Section \ref{sec:conclusions} conclusions about the work are drawn.

\section{Related Work}
\label{sec:related_work}

\textbf{Image Style Transfer.} RGB-to-IR CMST is a case of image style transfer(ST), which has been explored extensively in the context of texture mapping \cite{Li2020, ashikhmin2003fast, lee2010directional}.  Modern ST methods rely primarily upon DNNs to align the distribution of the stylized imagery with real-world imagery in the target style. In a seminal work Gatys et. al.  \cite{gatys2016image}, the authors hypothesized that the semantic content of an image and the style of an image (i.e., colors and textures) could be independently extracted from different activations in a DNN. Content features (from deep layers) and style features (from shallow layers) are extracted from two images $I_c$ and $I_s$ respectively. The output stylized image $I$ is iteratively constructed and has the semantic content from $I_c$ while emulating the style of $I_s$. 

Following the work in \cite{gatys2016image}, two general communities of ST emerged \cite{Li2020}. The first method we will discuss builds upon the approach in \cite{gatys2016image} directly and uses DNNs to isolate and align style and content from imagery. We refer to these types of methods as photorealistic style transfer (PST). The work in \cite{li2017universal} proposed the whitening coloring transform (WCT) 
that allows for non-unique optimization of $I_c$ and $I_s$, increasing the efficiency of stylization for large batches of images. 
WCT2 \cite{yoo2019photorealistic} introduces a novel encoder with lossless pooling layers and uses multiple WCT transforms in their encoder for a more robust stylization. In this work, we use WCT2 as a representative example of PST models due to its state-of-the-art performance compared to other PST models \cite{yoo2019photorealistic}.

The second general class of ST methods is termed adversarial style transfer (AST).  In these approaches some model (typically a DNN) - termed the critic or discriminator - is trained to distinguish stylized imagery from real imagery. Concurrently, the stylization model is trained to maximize the error of the critic.  
The cycle-consistent generative adversarial network (CycleGAN) \cite{zhu2017unpaired} is one widely used AST model \cite{Li2020, jing2019neural}. However, a known problem with CycleGAN is that it can erroneously alter the semantic content of images \cite{zhu2017unpaired}, e.g. warp or remove objects in the imagery. We provide an example of this occurring in the supplemental. CyCADA \cite{hoffman2018cycada} addresses this limitation by adding a semantic consistency objective to the model as well as a supervised task objective (e.g., classification, segmentation, detection, etc.). In this work, we use CyCADA as a representative example of AST models based on its state-of-the-art performance.

\textbf{RGB-to-IR CMST.} RGB-to-IR CMST can be conceived as a special case of ST where the goal is to create synthetic IR using RGB imagery as input. Some work in the  CMST community uses AST methods directly such as \cite{jameel2022generating}, and \cite{Cygert2019} who applied adversarial auto-encoders and CycleGAN to RGB-to-IR CMST respectively. Other work builds upon more general ST models. In \cite{Liu_2018_CVPR_Workshops} the authors add a content-preserving connection structure to the generator of CycleGAN, and an ROI loss objective. Liu et. al. \cite{Liu_2018_CVPR_Workshops} add a supervised task loss to CycleGAN in the form of object level classification. The task loss incentivizes class conditional stylization, as objects of different classes/materials have different thermal characteristics and provides additional information during training \cite{Liu_2018_CVPR_Workshops}. The authors in \cite{Kieu2020} replace the adversarial objective with a distance and add a perceptual loss. While neither \cite{Liu_2018_CVPR_Workshops} or \cite{Kieu2020} have made their code publicly available, they are conceptually similar to CyCADA.

A subset of AST models assumes the availability of paired RGB-IR imagery \cite{Li2020, isola2017image}. Pix2Pix \cite{isola2017image,abbott2020unsupervised}, ThermalGAN \cite{kniaz2018thermalgan}, and InfraGAN \cite{ozkanouglu2022infragan} have  been used for RGB-to-IR CMST.  RGB and IR image pairs provide a direct mapping of pixel intensities from the RGB to the IR style, however paired imagery is not always available for IR detection problems. Although our proposed approach, MLST, does not require paired imagery, we also compare to ThermalGAN and InfraGAN as state-of-the-art examples of paired ST models.  

Another strategy has been to hand-craft transformations based upon physics or heuristics \cite{Herrmann2019}. The authors in ThermalDet \cite{Herrmann2019} hand-crafted an effective CMST by composing several common image transformations (e.g., contrast enhancement, blurring) \cite{Buslaev2020}. We include ThermalDet \cite{Herrmann2019} in our benchmarking result as a strong hand-crafted baseline.  Although not included in previous studies, we also include several simple hand-crafted transformations as baselines (e.g., identity, grayscale transforms). 
 
As discussed in Section \ref{sec:introduction}, one limitation of existing work is that there is yet to be a comprehensive comparison of RGB-to-IR CMST approaches. Existing studies typically only compare their proposed methods against just one or two existing methods, and they do so on a limited number of datasets \cite{Herrmann2019, jameel2022generating, li2019multi, Cygert2019}. In this work we compare to numerous methods, including existing state-of-the-art CMST methods, representative recent methods from the general ST community, and simple hand-crafted methods. We compare these methods on four public benchmark datasets - more than any previous study on this topic: FLIR \cite{flirdataset}, DSIAC \cite{dsiacdataset}, CAMEL \cite{Gebhardt2019}, and Kaist \cite{Choi2018}.  These datasets have been previously employed for RBG-to-IR CMST \cite{Kieu2020,Liu_2018_CVPR_Workshops,ozkanouglu2022infragan} as well as IR object detection \cite{devaguptapu2019borrow,abbott2020unsupervised},  \cite{yadav2020cnn}. A general overview of each dataset, summary statistics describing them, and any modifications that we made (where applicable) can be found in our supplemental materials.

\textbf{Meta Learning.} Meta learning is a field of ML focused on optimizing the learning process itself \cite{hospedales2020meta}. Methods involving meta learning are diverse and applied in a number of problem settings such as few-shot learning \cite{li2017meta},  reinforcement learning \cite{cubuk2018autoaugment}, and automatic augmentation  \cite{hataya2020faster}.

\vspace{-.5em}
\section{Problem Setting}
\vspace{-.25em}
\label{sec:problem_setting}
We are specifically concerned with the problem of unpaired RGB-to-IR CMST, wherein we assume that we have some collection of labeled RGB imagery, $(X^{S},Y^{S}) = \{(x_{i}^{S},y_{i}^{S})\}_{i=1}^{N^{S}}$, termed the \textit{source data}, where $y^{S} \in \mathcal{Y}$ may take different forms depending upon the recognition task of interest (e.g., segmentation, object detect, and classification). We assume that  $x_{i}^{S} \sim d^{S}$, where $d^{S}$ is termed the source domain distribution. We also assume the availability of some IR imagery of the form $(X^{T},Y^{T}) = \{(x_{i}^{T},y_{i}^{T})\}_{i=1}^{N^{T}}$ where $y^T \in Y$. Here we assume $x^T_i \sim d^T$ and that $d^T \neq d^S$. Our goal is to train a model of the form $y = f\theta(x)$ using both the source and target domain datasets. Where $\hat{x}^T = f(x^S)$ such that $\hat{x}^T \sim d^T$.

\textbf{Performance Evaluation.} Evaluating the quality of a stylization model (e.g., compared to other competing models) is challenging because there are generally multiple valid IR pairs that could be associated with a given RGB image, due to unknown factors such as weather, scene material composition, or camera properties. This makes it difficult to evaluate the efficacy of a stylization model even when we have paired RGB-IR data: i.e., an RGB and IR image of the same scene, collected at the same time. To overcome this obstacle, we evaluate CMST models based on the effectiveness of its imagery when training a downstream task model of the form $y^{T} = f_{\gamma}(x^{T})$, following recent studies on RGB-to-IR stylization \cite{Kieu2020, Liu_2018_CVPR_Workshops, kniaz2018thermalgan, ozkanouglu2022infragan}.  In our experiments we specifically utilize object detection.   

\textbf{Supervised vs. unsupervised CMST} Existing CMST models are often designed with specific assumptions about the data that is available for training $\hat{g}_{\theta}$, resulting in several specific sub-problems of CMST. One potential assumption regards the availability of task-specific labels in the source and/or target domains (i.e., $Y^{S}$ and $Y^{T}$).  If such labels are assumed available, then we refer to this as \textit{supervised} CMST, and \textit{unsupervised} CMST otherwise.

\vspace{-.25em}
\section{Meta Learning Style Transfer}
\label{sec:mlst_description}

In this section we describe our proposed approach, termed Meta Learning Style Transfer (MLST), which models RGB-to-IR CMST as a composition of functions (e.g., contrast changes, blur) drawn from some fixed set of candidates, termed the dictionary, as illustrated in Fig. \ref{fig:mlst_concept}. The results of our experiments in Sec. \ref{sec:experiments} indicated that CMST models comprising a hand-chosen composition of a few simple transformations often outperformed more expressive DNN-based approaches (e.g., WCT2, CycleGAN). We hypothesize that these hand-crafted compositional methods might be made more effective if their design was data-driven rather than manual.  MLST is based upon Faster AutoAugment \cite{hataya2020faster}, an approach for learning effective image augmentation strategies; motivated by the benchmark results above, we adapt it to the task of image stylization and investigate its effectiveness for this different task. 

\textbf{Model Description.} In MLST we assume the availability of a set, or Dictionary, of $N$ candidate functions, where $O^{(n)}(\cdot)$ represents the $nth$ function in the Dictionary with $n \in [1,2,...N] $. Each function has associated with it a parameter (i.e., a magnitude or a threshold) $\mu^{(n)}$, and a a parameter for a probability distribution $p^{(n)}$ that describes the probability of that function being applied in \ref{eq:operation_definition}, where $p \in [0,1]$. This parameter $p$ is to account for variability within IR imagery. Each function is represented by a Bernoulli random variable of the form

\begin{equation}
\label{eq:operation_definition}
    x \leftarrow 
    \Biggl\{
     \begin{matrix}
        O^n(x; \mu^{(n)}) & p= p^{(n)}\\
        x & p=1-p^{(n)} \\
    \end{matrix} 
\end{equation}

In principle the user can include any function whose input and output are both images. However, in our implementation we choose commonly used image transformation functions that could contribute towards a physically plausible RGB-to-IR stylization (e.g., change in image contrast) and exclude functions that are not physically plausible (e.g., color channel shuffle). We limit the functions in this manner to simplify the search space for a good combination of functions. 

We define a Stylization Policy as a composition of $K$ functions applied in series where $K$ is a hyperparameter specifying the max number of applied functions.

\vspace{-.25em}
\begin{algorithm}
\caption{MLST Stylization}
\label{alg:mlst_inference}
\begin{algorithmic}

\State \textbf{Inputs}: 
\State x: Image
\State $O$: a function dictionary  where $O_k^{(n)}$ represents the nth function in the dictionary at the $kth$ step in the policy
\State Learned Parameters: $w_k^{(n)}$,  $\mu_k^{(n)}$, $p_k^{(n)}$
\State \textbf{Output}: Transformed Image, x 
\State 

\While {$k$ in {1, ... K}}
\State select function to apply: $(n) \sim \mathcal{C}[\sigma(w_k)]$
\State p $\sim$ Bern($p_k^{n}$)
\State if(rand() $\leq$ p): 
\State \hspace{10pt}  apply function to data: $x \leftarrow  O_k^{(n)}( x; \mu_k^{(n)},\mu_k^{(n)})) $
\State else: 
\State \hspace{10pt}  $x \leftarrow x$

\State $k++$
\EndWhile
\State return x

\end{algorithmic}
\end{algorithm}

\textbf{Model Parameter Optimization.} The goal of MLST is to estimate the parameters $w_k$ and $\mu_k$. We aim to develop a method that could utilize data to infer both (i) the functions that should be included in the composition, (ii) the order in which they should be applied, and (iii) any parameters $\mu$ associated with each function. For each $k$ step in the style transfer, unique $w_k$ and $\mu_k$ are found such that each step is independent of the prior step but includes the context of the order in which it is applied. During training, the output of a single operation is approximated as a weighted combination of all operations: 

\vspace{-.25em}
\begin{equation}    
x \leftarrow  \sum\limits_{n=1}^{N} [\sigma(w_k)]^{(n)} O_K^{(n)}( x; \mu_k^{(n)}, p_k^{(n)}),
\end{equation}

\noindent
where $\sigma$ is the softmax function and $w_k$ correspond to weights of a neural network. To find the ideal Stylization Policy, we differentiate $w_k$ with respect to $\mu$ and $p$. The goal is to learn $w_k$ such that $\sigma(w_k)$ behaves as a one-hot vector. During stylization (Algorithm \ref{alg:mlst_inference}), what function to apply to an image is sampled from a categorical distribution $\mathcal{C}([\sigma(w_k)]$. Direct optimization of this objective is not normally feasible due to the functions being non-differentiable. Such non-differentiable problems are the purview of Meta learning, which is a field of ML focused on optimizing the learning process itself \cite{hospedales2020meta}, such as reinforcement learning \cite{cubuk2018autoaugment}. \cite{hataya2020faster} implemented a reinforcement-learning like optimization structure in their work  Faster auto-augment for learning an ideal way to augment RGB imagery.  This is a different, and potentially simpler, operation than mapping from one modality to another. We adapted this method as the basis MLST.

For a critic, we used a ResNet-18 \cite{he2016deep} and replaced the classification layers with a two-layer network. A similar critic is used in \cite{hataya2020faster, cubuk2018autoaugment}. A classification ‘task’ loss is added to prevent images of a specific class from being stylized into images of another class. This approach was chosen to preserve the different spectral statistics associated with different materials which correspond to each class. For instance, people are usually bright in IR imagery but a car with the engine off is not bright. The training of our algorithm is shown in Algorithm  \ref{alg:mlst_train}.

\vspace{-.5em}
\begin{algorithm}
\caption{MLST Train}
\label{alg:mlst_train}
\begin{algorithmic}

\State \textbf{Inputs}: 
\State $f_\theta$: learned stylization model
\State $f_\gamma$: task model (adversarial critic)
\State $\epsilon:$ task loss coefficient
\State $\{X^T, Y^T\}$ , $\{X^S, Y^S\}$: training data sampled from IR and RGB domains respectively

\State
\While {not converged}

\State Sample a pair of batches $B^T, B^S$ from $X^T, X^S$
\State Stylize data $\hat{B}^T=f_\theta(x); (x \in B^S); \hat{Y}^T=Y^S$
\State Compute Wasserstein distance loss  $ \mathcal{L}_d = d(B^T, \hat{B}^T)$
\State Compute task loss $\mathcal{L}_\gamma=\sum\limits^{B_T}f_\gamma({x}^T,{y}^T) + \sum\limits^{\hat{B}^T}f_\gamma(\hat{x}^T,\hat{y}^T) $  
\State Compute total loss $\mathcal{L} = \mathcal{L}_d+ \epsilon \mathcal{L}_\gamma$
\State $\underset{w,\mu,p }{\argmin} (\mathcal{L})$
\EndWhile
\end{algorithmic}
\end{algorithm}
\vspace{-.5em}

\begin{figure*}[tp]
\centering
\includegraphics[width=0.85\textwidth]{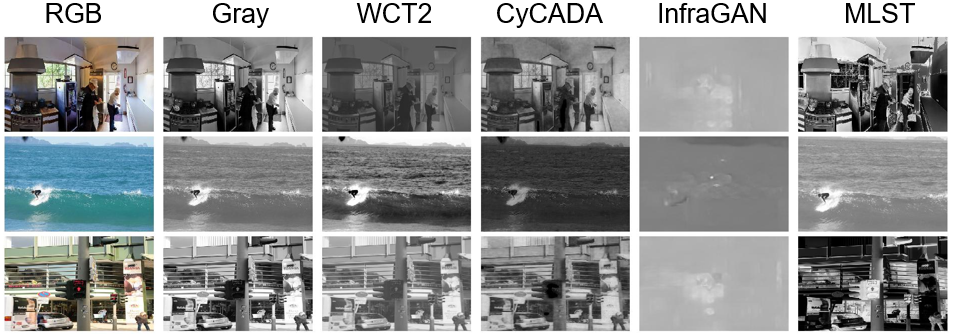}
\vspace{-0.70em}
\caption{Example imagery produced by each CMST method we evaluated. Each column of the matrix has five representative images all stylized by an NST algorithm. Gray and WCT2 are both very similar, CyCADA introduces artifacts into the imagery, and InfraGAN saturates images when deployed on out-of-domain images. MLST introduces no artifacts and produces a relatively complex nonlinear stylization.}
\label{fig:example_images}
\end{figure*}

\setlength{\tabcolsep}{24pt}
\begin{table*}[tp]
\normalsize
    \centering
    \caption{This table shows the results of our experiment, 
    comparing the detection performance of a YOLO model across each benchmark dataset; the training imagery for the model is supplemented with imagery generated from different CMST methods across each benchmark dataset. The winner in each dataset is bolded. The average accuracy across all datasets is included to summarize general performance. }
\vspace{-0.75em}
\label{tab:results}

\scalebox{.70}{
\begin{tabular}{cccccccc}
\hline
\hline
\rowcolor[HTML]{FFFFFF} 
\cellcolor[HTML]{FFFFFF}  & \cellcolor[HTML]{FFFFFF}  & \cellcolor[HTML]{FFFFFF} & \multicolumn{4}{c}{\cellcolor[HTML]{FFFFFF}\textbf{Benchmark Dataset Performance (mAP@0.5)}} & \cellcolor[HTML]{FFFFFF}                               \\ \cline{4-7}
\rowcolor[HTML]{FFFFFF} 
\multirow{-2}{*}{\cellcolor[HTML]{FFFFFF}\textbf{Method}} & \multirow{-2}{*}{\cellcolor[HTML]{FFFFFF}\textbf{Paired?}} & \multirow{-2}{*}{\cellcolor[HTML]{FFFFFF}\textbf{Supervised?}} & \textbf{FLIR}        & \textbf{DSIAC}        & \textbf{CAMEL}  & \textbf{KAIST} & \multirow{-2}{*}{\cellcolor[HTML]{FFFFFF}\textbf{AVG}} \\ 
\hline
\hline
\rowcolor[HTML]{EFEFEF} 
None   & N/A  & N/A  & .647 & .593 & .582 & .427 & .562 \\
\rowcolor[HTML]{FFFFFF} 
Identity  & N/A  & N/A  & .650 & .660  & .610   & .478    & .599  \\
\rowcolor[HTML]{EFEFEF} 
Grayscale (Inversion) & N/A  & N/A & \textbf{.676} & .660 & .684 & .488 & .627   \\
\rowcolor[HTML]{FFFFFF} 
Grayscale     & N/A & N/A & .666 & .640 & .693 & .478 & .619 \\
\rowcolor[HTML]{EFEFEF} 
ThermalDet     & N/A & N/A & .628 & .689 & .641 & .485 & .612 \\
\rowcolor[HTML]{FFFFFF} 
WCT2    & N & N & .642 & .660 & .643 & \textbf{.490} & .609 \\
\rowcolor[HTML]{EFEFEF} 
CycleGAN      & N & N & .600 & .669 & .603 & .451 & .581 \\
\rowcolor[HTML]{FFFFFF} 
CyCADA      & N & N & .640 & .684 & .634 & .469 & .608 \\
\rowcolor[HTML]{EFEFEF} 
CyCADA      & N & Y & .647 & .686 & .666 & .469 & .617 \\
\rowcolor[HTML]{FFFFFF} 
ThermalGAN  & Y & Y & .642 & .675 & .615 & .475 & .601 \\
\rowcolor[HTML]{EFEFEF} 
InfraGAN    & Y & N & .666 & .687 & .653 & .479 & .621 \\
\rowcolor[HTML]{FFFFFF} 
MLST (Ours)   & N & N & .629 & .716 & .749 & .459 & .638 \\
\rowcolor[HTML]{EFEFEF} 
\textbf{MLST (Ours)}   & N & Y & .653 & \textbf{.723} & \textbf{.751} & .471 & \textbf{.650} 
\\ \hline
\hline
\end{tabular} }

\end{table*}
\setlength{\tabcolsep}{6pt}

\section{Benchmark Stylization Models}
\label{sec:baseline_methods}
In our benchmark we include state-of-the-art models representing each of several classes of stylization methods. We first include some simple baselines: the \textit{None} method, which involves no supplementation with stylized imagery; the \textit{Identity} method, which does no stylization to the RGB imagery; \textit{Grayscale} which converts the RGB imagery to grayscale; \textit{Grayscale (Inversion)} which converts the imagery to grayscale, and then for each intensity, denoted $I$, it computes a new intensity, denoted $I'$ using the following transformation: $I'=255-I$, where we assume $I \in [0,255]$. \textit{ThermalDet} \cite{Herrmann2019} is a recent method comprising a hand-crafted composition of simple operations (e.g., grayscale) that was found highly effective. \textit{WCT2} \cite{yoo2019photorealistic} is used as a state-of-the-art example of PNST models. \textit{CycleGAN} \cite{zhu2017unpaired} and \textit{CyCADA} \cite{hoffman2018cycada} are used as representative examples of state-of-the-art adversarial style transfer models in this work. \textit{ThermalGAN} \cite{kniaz2018thermalgan} and \textit{InfraGAN} \cite{ozkanouglu2022infragan} are recent stylization methods specifically designed for RGB-to-IR CMST.  Notably, ThermalGAN and InfraGAN both require paired RGB-IR imagery, which is not always available, and it is only available in three of our four benchmarks. And ThermalGAN requires thermal segmentation labels for training, which is not available in any of our public benchmarks.  Therefore, we needed to make some adaptations of the training procedures for ThermalGAN and InfraGAN.  These additional details, as well as precise implementation details of each benchmark approach are provided in the supplemental materials. Importantly however, any model parameters that were not specified by the source publications (e.g., learning rates for each of our datasets) were optimized on our downstream detection task using a greedy forward sequential search, in order to help ensure a fair comparison. 

\section{Benchmark Datasets}
\label{sec:experimental_datasets}
\textbf{IR Datasets.} We sought to include IR datasets that labels to support the training of object detection models (for scoring stylization), and that are also publicly-available, to support future replication.  Using this criteria, we included four datasets: FLIR \cite{flirdataset}, DSIAC \cite{dsiacdataset}, CAMEL \cite{Gebhardt2019}, and KAIST, \cite{Choi2018}.  Additional details for each dataset are included in the supplement.  \textbf{RGB Datasets.}  To train our stylization models we also require RGB imagery.  For this purpose, we utilize the MS COCO dataset \cite{lin2014microsoft} because it has object detection labels, and because training upon it results in highly accurate object detection models for RGB.  We reason therefore that any accurate RGB-to-IR stylization of COCO imagery will result in imagery that is useful for training accurate IR object detectors; this is important since we use downstream IR object detection accuracy as a scoring metric for our RGB-to-IR stylization models.  Additionally, we use MS COCO because it is publicly-available and widely-used.

\vspace{-.5em}
\section{Experiments}
\vspace{-.25em}
\label{sec:experiments}

We conduct experiments to study the efficacy of each CMST method described in Sec. \ref{sec:baseline_methods}, as well as our proposed MLST method. To measure the efficacy of each stylization model, we evaluate the performance of IR object detection models when they are trained on stylized RGB imagery (i.e., synthetic IR imagery) produced by that model.  

\textbf{Stylization Model Training.}  We train the $p^{th}$ stylization model, denoted $f_{\theta}^{(p)}$, using two sets of imagery: the training partition of the MS COCO dataset, denoted $D_{RGB}$, and an IR dataset comprising the training partition from each of our four benchmark IR datasets in Sec. \ref{sec:experimental_datasets}, denoted $D_{IR}$.  For each of the aforementioned datasets we use previously-established training partitions (see Supplemental), and every stylization model is trained on exactly the same dataset. 

\textbf{Performance Evaluation of Stylization.} Once each model is trained, we use it to stylize the imagery in $D_{RGB}$ to create a new dataset of synthetic IR, denoted $D_{RGB \rightarrow IR}(p)$, where $p$ indicates that it was stylized by the $p^{th}$ model. For each benchmark IR dataset in Sec. $\ref{sec:experimental_datasets}$, we combine $D_{RGB \rightarrow IR}(p)$ with the training partition of that dataset.  Let $D_{IR}(t)$ indicate the training subset of $D_{IR}$ corresponding to the $t^{th}$ IR benchmark dataset from Sec \ref{sec:experimental_datasets}, then for each combination of stylization model and benchmark dataset, we create a training dataset $D_{train}(p,t) = D_{RGB \rightarrow IR}(p) \cup D_{IR}(t)$.  We then use $D_{train}(p,t)$ to train an object detection model and evaluate its accuracy on the testing partition of the $t^{th}$ benchmark dataset.  Following convention in IR object detection (e.g., \cite{flirdataset,Choi2018}), we evaluate the mean average precision at an IoU of 0.5, or mAP@0.5 (mAP) score, for the trained detector.  We repeat this process for all combinations of stylization model, $p$, and benchmark dataset, $t$.  

\textbf{Task Model and Training} For the task model, we use a YoloV3 (YOLO) due to its wide use in the literature, with the implementation described in \cite{Jocher}. YOLO models were trained to convergence on this stylized COCO dataset and the combined training data from each benchmark dataset. We observed these models converged in about 40 epochs for each case.  Because stylized RGB imagery is usually lower-quality than real IR imagery, we limit its influence on the training process by constructing each minibatch of 8 comprising one stylized RGB image, and 7 real IR images. 

\textbf{Optimization of stylization hyperparameters.} To help ensure a fair comparison of stylization approaches, we optimized important data-dependent hyperparameters for each model (e.g., learning rate) using a greedy sequential forward search, where we choose hyperparameters with the highest accuracy on the downstream object detection task described above, and using a validation dataset that is disjoint from the testing set.  

\vspace{-.25em}
\subsection{Experimental results}
\vspace{-.125em}
\label{subsec:experimental_results}

The quantitative results of our experiments are shown in Table \ref{tab:results}. The results indicate that every approach considered is beneficial (i.e., higher mAP score, on average) compared to using no supplementation i.e., the \textit{None} method. In the \textit{None} method no extra data was used, and the model was optimized and finetuned on the IR only. Surprisingly, even using the \textit{Identity} stylization - equivalent to supplementing with raw RGB imagery - is still highly beneficial.  Furthermore, this approach is consistently beneficial; on all four benchmark datasets it improves over \textit{None}.  We hypothesize that the diversity of content in the MS COCO dataset outweighs the disadvantages of training directly upon RGB imagery, which is a different modality and therefore known to be visually distinct compared to IR a priori.  

The \textit{Identity} method also represents an important baseline for any other more sophisticated CMST methods; if they are effectively stylizing the RGB imagery then we would expect that these methods should be superior to \textit{Identity}.  The results indicate that simple \textit{Grayscale (Inversion)} and \textit{Grayscale} methods do indeed improve over the naive \textit{Identity} method. These results suggest that, despite their simplicity, these methods do \textit{tend} to make RGB imagery more visually similar to IR. Notably, the Grayscale(Inversion) is never harmful compared to \textit{Identity}, and improves over it in all four benchmark datasets. 

Surprisingly, nearly all of the data-driven CMST models perform more poorly (i.e., lower mAP) than the simple \textit{Grayscale (Inversion)} method. The only exception to this trend is our proposed approach, \textit{MLST}. Example stylized imagery produced by several of the data driven models is shown in Fig. \ref{fig:example_images}. Based upon this imagery it appears that the data-driven models sometimes create artifacts in the stylized imagery, or reduce the dynamic range of the pixel intensities, which we hypothesize was ultimately detrimental to task model performance. By contrast, and by design of its function dictionary, MLST introduces few or no artifacts into the imagery, and seems to maintain or expand the dynamic range of the stylized imagery.  

The best performing CMST model on average is MLST, both in a supervised and unsupervised setting. As expected, the supervised MLST model achieves somewhat better performance, owing to the use of additional information during training (i.e., downstream task labels).  We also observe that supervision improved \textit{CyCADA}, although it achieved lower overall performance. The unsupervised and supervised MLST models improved over the \textit{Baseline} method by $13.5$\% and $15.7$\% respectively. Importantly, they also improve over the simple Grayscale (Inversion) method, on average, and are the only two data-driven methods to do so. Although supervised MLST does not always achieve the best performance on every individual dataset, it achieves the best performance compared to all other models on two of the four benchmark datasets, and the best average performance across all datasets.

\setlength{\tabcolsep}{6.5pt}
\begin{table}[tp]
\normalsize
    \centering
    \caption{This table shows the results of a study described in \ref{sec:analysis}, comparing the detection performance of a YOLO model across each benchmark dataset; the training imagery for the model is supplemented with stylized imagery} 
\vspace{-0.75em}
\label{tab:style_vs_aug}
\scalebox{.70}{
\begin{tabular}{ccccccc}
\hline
\hline
\rowcolor[HTML]{FFFFFF} 
\cellcolor[HTML]{FFFFFF} &  \textbf{Uses} & \multicolumn{4}{c}{\cellcolor[HTML]{FFFFFF}\textbf{Performance (mAP@0.5)}} \\ 
\cline{3-6}
\rowcolor[HTML]{FFFFFF} 
\multirow{-2}{*}{\cellcolor[HTML]{FFFFFF}\textbf{Method}} & \textbf{COCO?} & \textbf{FLIR}        & \textbf{DSIAC}        & \textbf{CAMEL}  & \textbf{KAIST} & \multirow{-2}{*}{\cellcolor[HTML]{FFFFFF}\textbf{AVG}} \\ 
\hline
\hline
\rowcolor[HTML]{EFEFEF} 
None   &  &  &  &  &  &    \\
\rowcolor[HTML]{EFEFEF} 
(Baseline)  & \multirow{-2}{*}{N}   & \multirow{-2}{*}{.647} & \multirow{-2}{*}{.593} & \multirow{-2}{*}{.582} & \multirow{-2}{*}{.427} & \multirow{-2}{*}{.562} \\
\rowcolor[HTML]{FFFFFF} 
Grayscale   &  &  &  &  &  &    \\
(Inversion) & \multirow{-2}{*}{Y}  & \multirow{-2}{*}{\textbf{.676}} & \multirow{-2}{*}{.660} & \multirow{-2}{*}{.684} & \multirow{-2}{*}{\textbf{.488}} & \multirow{-2}{*}{.627}   \\
\rowcolor[HTML]{EFEFEF} 
Aug     & N & .632 & .657 & .641 & .419 & .587 \\
\rowcolor[HTML]{FFFFFF} 
MLST Aug & Y & .646 & .664 & .610 & .454 & .594 \\
\rowcolor[HTML]{EFEFEF} 
MLST Style   & Y & .653 & \textbf{.723} & .751 & .471 & .650 \\
\rowcolor[HTML]{FFFFFF} 
MLST Style +   &  &  &  &  &  &    \\
MLST Aug  & \multirow{-2}{*}{Y} & \multirow{-2}{*}{.662} & \multirow{-2}{*}{.713} & \multirow{-2}{*}{\textbf{.769}} & \multirow{-2}{*}{.481} & \multirow{-2}{*}{\textbf{.654}} \\
\hline
\hline
\end{tabular} }
\vspace{-0.75em}
\end{table}
\setlength{\tabcolsep}{6pt}

\vspace{-.5em}
\section{Additional Analysis}
\vspace{-.25em}
\label{sec:analysis}

\begin{figure}[tp]
\centering
\includegraphics[width=0.99\columnwidth]{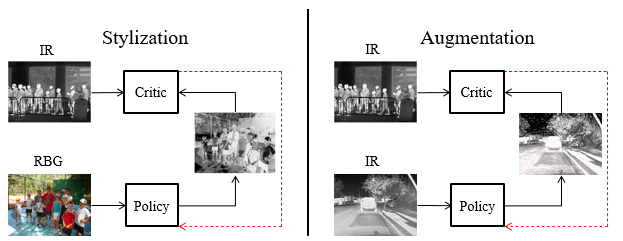}
\vspace{-0.75em}
\caption{This figure shows how MLST is trained for stylization (top) and how MLST is trained for augmentation (bottom). For stylization only RGB imagery is acted upon by the policy during training. For augmentation IR imagery is acted upon by the policy in training.}

\label{fig:style_vs_aug}
\vspace{-1.em}
\end{figure}

\textbf{Do We Really Need the RGB Imagery?} In this section we consider whether we could achieve the same results with MLST by simply using it to learn to \textit{augment} the existing IR datasets, rather than stylize IR imagery.  We answer this question by learning one MLST policy for stylization (i.e., RGB-to-IR), and one for augmentation (i.e., IR-to-IR), as illustrated in Fig. \ref{fig:style_vs_aug}. We also train an MLST model to perform both augmentation and stylization, as these may be complementary. The results of our experiment are shown in Table \ref{tab:style_vs_aug}. Both augmentation and stylization policies yield better performance for a task model on all benchmark datasets. However, MLST for stylization yields a substantially larger mAP score improvement on our detection task than MLST for augmentation. This suggests that the benefits of MLST are primarily due to effective stylization of the RGB imagery. Additionally, stylization and augmentation are additively beneficial, the model trained with both policies has an even higher mAP score. 

\textbf{What Does MLST Learn To Do?} We have observed that the stylization and augmentation policies lead to different results in Table \ref{tab:style_vs_aug}. We want to consider if the training objectives of stylization and augmentation change how MLST learns to utilize the function dictionary. Do the two policies use significantly different functions? This would suggest the policies are performing distinct operations.  To aid in the understanding of stylization policies learned by our model, it is helpful to compute two summary statistics: the expected number of times each operation is applied to an image

\begin{equation}
\vspace{-.5em}
\label{eq: expected_p}
   \mathit{\mathbb{E}}[T^{(n)}] \approx \sum\limits_{k} \sigma(w_k)^{(n)}
\end{equation}

\noindent and the expected parameter value of an operation

\begin{equation}
\vspace{-.25em}
\label{eq: expected_mu}
    \mathit{\mathbb{E}}[\mu^{(n)}] \approx \sum\limits_{k} \sigma(w_k)^{(n)} \mu_k^{(n)}
\end{equation}

\begin{figure}[tp]
\centering
\includegraphics[width=0.99\columnwidth]{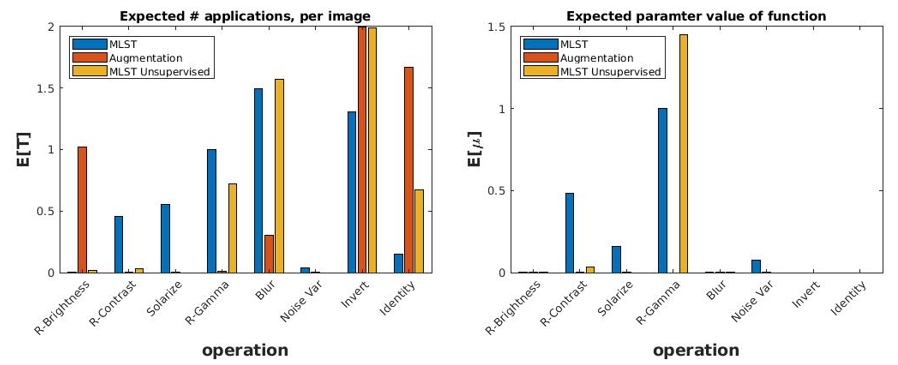}
\vspace{-1em}
\caption{This figure shows the differences in learned policies of MLST when used for stylization (blue), augmentation (orange), and unsupervised MLST (yellow). The top plot shows the expected number of times each operation in the dictionary is applied to an image $E[T^n]$ The lower plot shows the average parameter value of each operation $E[\mu^{(n)]}$.  Note that operations invert and identity have no parameter and their $E[\mu^{(n)]}$ is set to zero.}

\label{fig:mlst_summary}
\vspace{-1.5em}
\end{figure}

A summary of the different policies learned is shown in Fig. \ref{fig:mlst_summary}, we also include the unsupervised MLST stylization policy from Sec. \ref{sec:experiments} to see if the lack of supervision significantly alters what functions MLST learns to use. We compare three policies: MLST, MSLT for augmentation, and unsupervised MLST. The results indicate that the stylization and augmentation policies are largely distinct; each policy employs the available operations at very different frequencies. The stylization uses a more diverse set of functions compared to the augmentation policy. The stylization policy learned to prioritize the use of 5 functions: Random contrast, solarize, gamma transform, blur, and inversion. The augmentation policy learned to prioritize inversion and a random brightness shift; this is similar to Grayscale (inversion) the best non data driven stylization in our experiments. Using UMAP \cite{McInnes2018} dimensionality reduction, we can inspect the distribution of image features directly, and we found evidence that the stylized images are closer in feature space to the IR images than their RGB counterparts. Further details and associated UMAP visualizations are provided in the supplemental materials.

\vspace{-.5em}
\section{Conclusions}
\vspace{-.25em}
\label{sec:conclusions}
In this work we conducted experiments to measure the effectiveness of RGB-to-IR stylization methods, as measured by the impact of supplementing training data for supervised IR object detection models. We investigated seven contemporary style transfer methods across four different publicly-available IR datasets. Our experiments indicated that data supplementation with stylized RGB imagery often yields substantial performance improvements, and that the most effective methods on average consisted of compositions of simple transformations (e.g., grayscale, inversion).  Inspired by these findings, we developed MLST, which automatically learns a useful composition of functions and their parameters and obtained the greatest average performance among all methods considered.  

\textbf{Limitations and Future Work.} Although successful, MLST is only proven successful for RGB-to-IR stylization.  This is primarily because MLST assumes the availability of a dictionary of appropriate functions for the task is available, which is true for RGB-to-IR, but is not generally true for stylization tasks. Future work may focus on directly addressing the limitations of more expressive stylization models (e.g., CycleGAN, Cycada, WCT2), which in principle could learn more accurate transformations, even though empirically they did not perform well for RGB-to-IR stylization on our benchmarks. Future work can also be done to extend the use of MLST to other image domains such as hyperspectral and synthetic aperture radar (SAR) images.

\vspace{-.5em}
\section*{Acknowledgment}
\vspace{-.25em}
This work was supported by the U.S. Army CCDC C5ISR RTI Directorate, via a Grant Administered by the Army Research Office under Grants W911NF-06-1-0357 and W911NF- 13-1-00.

{\small
\bibliographystyle{ieee_fullname}
\bibliography{Main}
}

\end{document}